\documentclass{article}
\usepackage{spconf,amsmath,graphicx,hyperref}

\usepackage{xcolor}
\usepackage{amssymb,amsfonts}
\usepackage{textcomp}
\usepackage{cuted}
\usepackage{caption}
\usepackage{enumitem}
\usepackage{booktabs}
\usepackage{algorithm}
\usepackage{algorithmic}
\usepackage{multirow}
\usepackage{multicol}

\title{Freq-DP Net: A Dual-Branch Network for Fence Removal using Dual-Pixel and Fourier Priors}
\name{Kunal Swami$^{\star}$, Sudha Velusamy$^{\star}$, Chandra Sekhar Seelamantula$^{\dagger}$}
\address{$^{\star}$ Visual Intelligence Team, Samsung Research India Bangalore\\ 
         $^{\dagger}$ Department of Electrical Engineering, Indian Institute of Science}
\begin{document}
%
\maketitle
\begin{abstract}
Removing fence occlusions from single images is a challenging task that degrades visual quality and limits downstream computer vision applications. Existing methods often fail on static scenes or require motion cues from multiple frames. To overcome these limitations, we introduce the first framework to leverage dual-pixel (DP) sensors for this problem. We propose Freq-DP Net, a novel dual-branch network that fuses two complementary priors: a geometric prior from defocus disparity, modeled using an explicit cost volume, and a structural prior of the fence's global pattern, learned via Fast Fourier Convolution (FFC). An attention mechanism intelligently merges these cues for highly accurate fence segmentation. To validate our approach, we build and release a diverse benchmark with different fence varieties. Experiments demonstrate that our method significantly outperforms strong general-purpose baselines, establishing a new state-of-the-art for single-image, DP-based fence removal.
\end{abstract}

\vspace{-0.2cm}
\section{Introduction}
\label{sec:intro}

In an era where computer vision systems demand clear visual data, real-world scenes are frequently compromised by foreground obstructions. Among the most pervasive are fences, whose diverse structures, varying in material, thickness, and pattern density, pose a formidable challenge. These occlusions create a critical failure point for automated systems, catastrophically degrading the performance of object detectors \cite{occlusionhandlingobjectdetection_cvpr2020}, segmentation models, and trackers essential for autonomous driving \cite{occlusion_automotives_tits2019} and robotics \cite{occlusion_robotics_iros2021}. Effectively ``seeing through" these varied fences from a single image is thus a crucial step towards robust machine perception.

\begin{figure}[t!]
	\centering
	\includegraphics[scale=0.24]{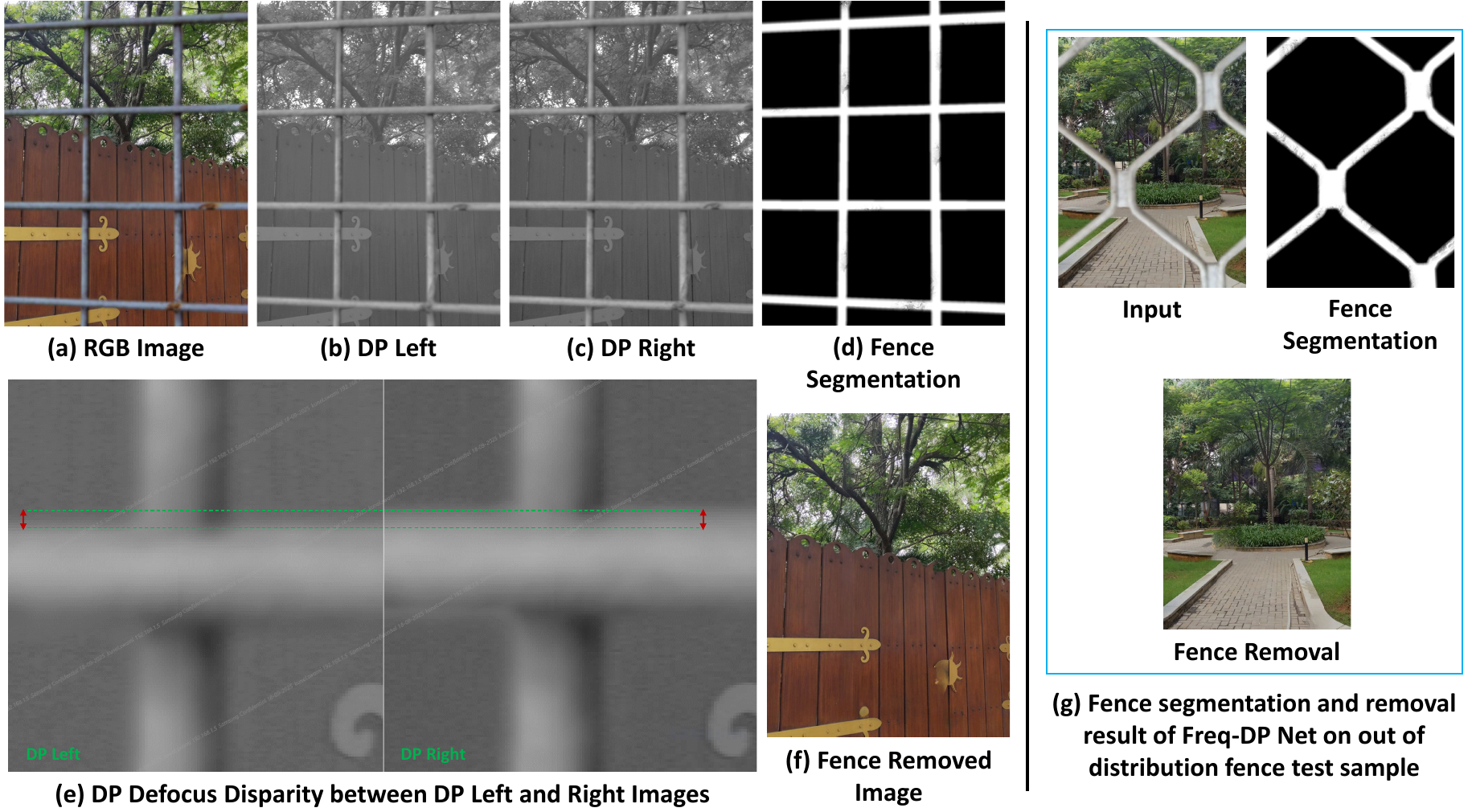}
    \vspace{-0.3cm}
	\caption{From a single input (a), we leverage disparity between the left (b) and right (c) DP views to generate a precise mask (d), which guides a model to produce the final result (f). Zoomed-in patches (e) highlight the disparity cue. DP views are grayscale (from the sensor's green channel). We use `left/right' terminology for consistency, the smartphone disparity is typically vertical. (g) shows our result on an out-of-distribution fence sample. Please zoom in for better detail.}
	\label{fig:teaser}
    \vspace{-0.6cm}
\end{figure}

Existing solutions for this problem face significant limitations. Methods that rely on motion parallax from multi-frame captures \cite{efficient_defencing_wacv2023,video_defencing_icme2018} are often impractical for the user and fail entirely in static scenes, while single-image approaches struggle with the inherent ambiguity between the fence and background. Furthermore, the development of truly generalizable solutions has been hampered by a narrow focus on limited, non-diverse datasets \cite{video_defencing_icme2018}, leading to methods that overfit to specific patterns. To overcome these challenges, we introduce the first framework to leverage the rich physical information from a dual-pixel (DP) sensor \cite{dp_exploration_cvpr2021,synthesize_dp_iccp2023}. Our approach is built on fusing two complementary cues: geometric information derived from the fence's defocus disparity, which relies on the practical assumption that the background is in focus, and the fence's global structural patterns captured in the frequency domain. This fusion of appearance-invariant priors allows us to achieve a new state-of-the-art in fence removal, as demonstrated in our teaser (Fig.~\ref{fig:teaser}), enabling more robust visual understanding.

Following are the \textbf{major contributions} of our method: 
\vspace{-0.2cm}
\begin{itemize}[leftmargin=*]
    \setlength{\itemsep}{0cm}
    \item The first framework for single-image fence removal using appearance-invariant disparity cues from a DP sensor.
    \item Freq-DP Net, a novel dual-branch network that fuses geometric cues from a DP cost volume with global structural patterns learned via Fast Fourier Convolutions (FFCs).
    \item We build and release the most diverse benchmark for this task, featuring varied fences and a test set of $150$ real-world (RGB-DP-GT) and $100$ synthetic (RGB-DP-GT-Mask) scenes.
    \item A new state-of-the-art on our benchmark, significantly outperforming strong general-purpose baselines.
\end{itemize}

\vspace{-0.38cm}
\section{Related Work}
\label{sec:relatedwork}

\vspace{-0.25cm}
\subsection{DP Sensor}
\label{subsec:dpsensor}
\vspace{-0.15cm}

\begin{figure}[t!]
	\centering
	\includegraphics[scale=0.82]{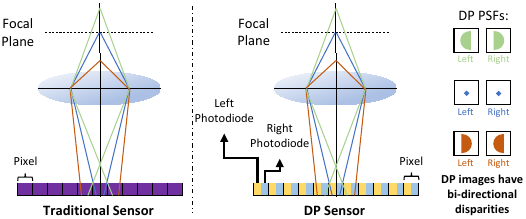}
    \vspace{-0.25cm}
	\caption{DP sensor image formation.}
	\label{fig:dualpixelsensor}
    \vspace{-0.45cm}
\end{figure}

A dual-pixel (DP) sensor captures two perspectives of a scene in a single shot by splitting each pixel into left and right photodiodes (Fig.~\ref{fig:dualpixelsensor}). While in-focus points are imaged identically, out-of-focus points project to different locations, creating a shift between the left-view ($I_L$) and right-view ($I_R$) known as defocus disparity. The disparity's direction depends on whether a point is in front of or behind the focal plane, and its magnitude is proportional to the point's distance from it. This relationship can be modeled by convolving a latent sharp image $I_{sharp}$ with distinct, depth-dependent Point Spread Functions (PSFs) for each view ($k_d^L, k_d^R$) \cite{synthesize_dp_iccp2023,dpdnet_eccv2020}:

\vspace{-0.3cm}
\begin{equation}
    I_v = I_{sharp} \circledast k_d^v, \quad v \in \{L, R\}
    \label{eq:dp_formation}
\end{equation}
\vspace{-0.5cm}

where the combined image $I_C = (I_L + I_R)/2$ is blurred by an effective PSF $k_d^C = (k_d^L + k_d^R)/2$. This defocus disparity provides a powerful, appearance-invariant cue for occlusion analysis. Note that smartphones limit DP photodiodes to the green pixels, producing grayscale DP views.

\vspace{-0.4cm}
\subsection{Fence Removal}
\label{subsec:fenceremoval}
\vspace{-0.15cm}

Prior works on obstruction removal often model the scene as a composite of two layers \cite{obstruction_free_photo_siggraph2015, sold_tpami2022}, typically leveraging motion parallax from video to distinguish them. This reliance on multi-frame captures is problematic as it is often impractical for the user, fails for static scenes, and in some cases, requires a pre-existing fence mask for removal \cite{obstruction_free_photo_siggraph2015,sold_tpami2022}.

Fence-specific literature has similarly evolved from classical methods \cite{img_defencing_cvpr2008, img_defencing_rev_accv2010} to deep learning approaches that depend on multi-frame bursts \cite{efficient_defencing_wacv2023}, video \cite{video_defencing_icme2018}, or multiple cameras \cite{stereo_defencing_icassp2017}. Progress in this area has been critically bottlenecked by the reliance on a single, non-diverse public dataset \cite{video_defencing_icme2018}, which contains only one type of metallic, diamond-patterned fence, leading to methods that overfit to a simple pattern and its appearance \cite{efficient_defencing_wacv2023}. \textcolor{black}{While recent efforts like Instruct2See \cite{instruct2see_icml2025} have begun addressing these cross-distribution challenges by learning to remove varied obstructions, significant gaps remain:} (i) the lack of a diverse benchmark for robust evaluation, and (ii) an over-reliance on motion or appearance cues, creating a need for a single-image method that uses a more robust physical prior.

\vspace{-0.3cm}
\section{Dataset Generation}
\label{sec:datasetgeneration}
\vspace{-0.15cm}

\begin{figure}[t!]
	\centering
	\includegraphics[scale=0.20]{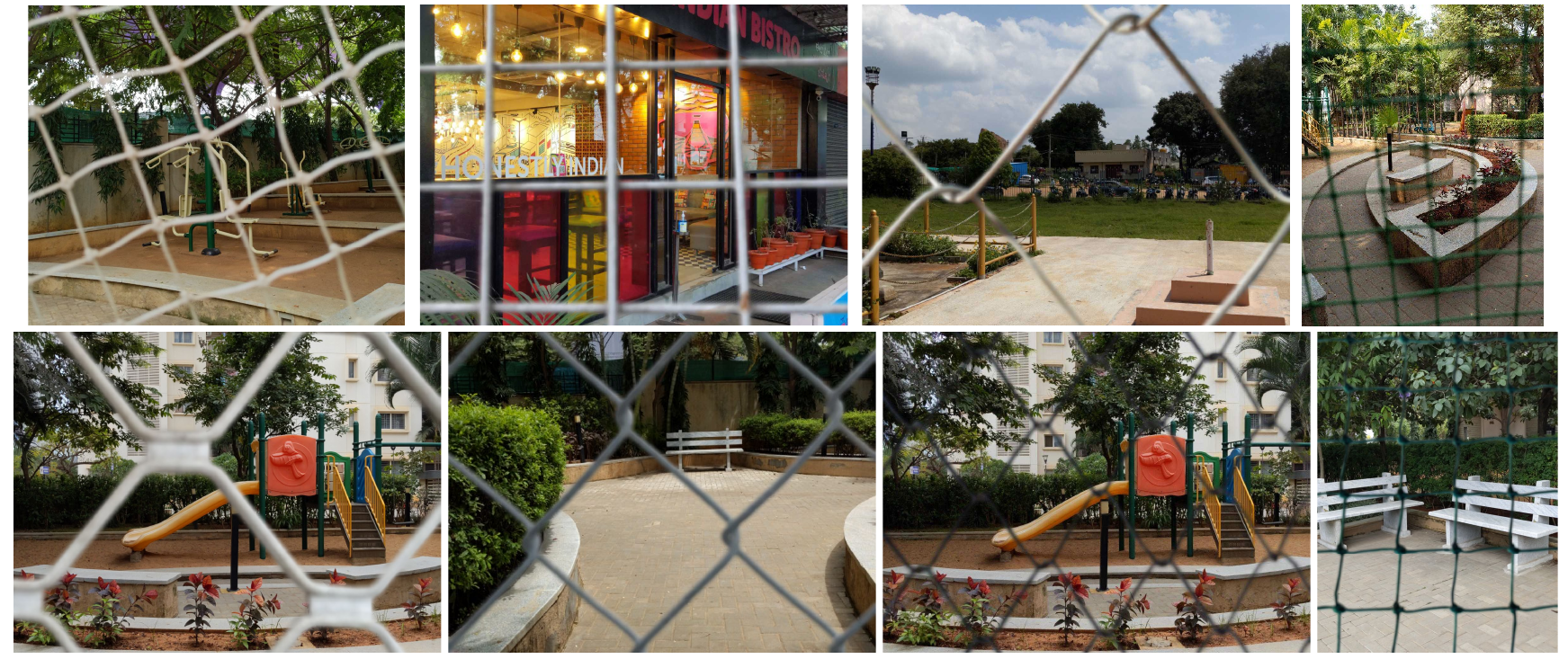}
    \vspace{-0.3cm}
	\caption{Our dataset features a wide range of fence structures, varying in pattern, material, and thickness.}
	\label{fig:diversityoffences}
    \vspace{-0.25cm}
\end{figure}

A core contribution of this work is the establishment of the first comprehensive framework for DP based fence removal, necessitated by the absence of public data for this task. We introduce two key components: (i) a synthetic data pipeline for generating a large-scale training set, and (ii) a public benchmark for evaluation. Unlike existing works that are limited to a single fence type \cite{video_defencing_icme2018}, our benchmark consists of both synthetic and captured real-world test scenes featuring a wide variety of fence structures (see Fig.~\ref{fig:diversityoffences}). This enables, for the first time, standardized training and rigorous comparison of methods on this challenging problem.

\begin{figure}[t!]
	\centering
	\includegraphics[scale=0.25]{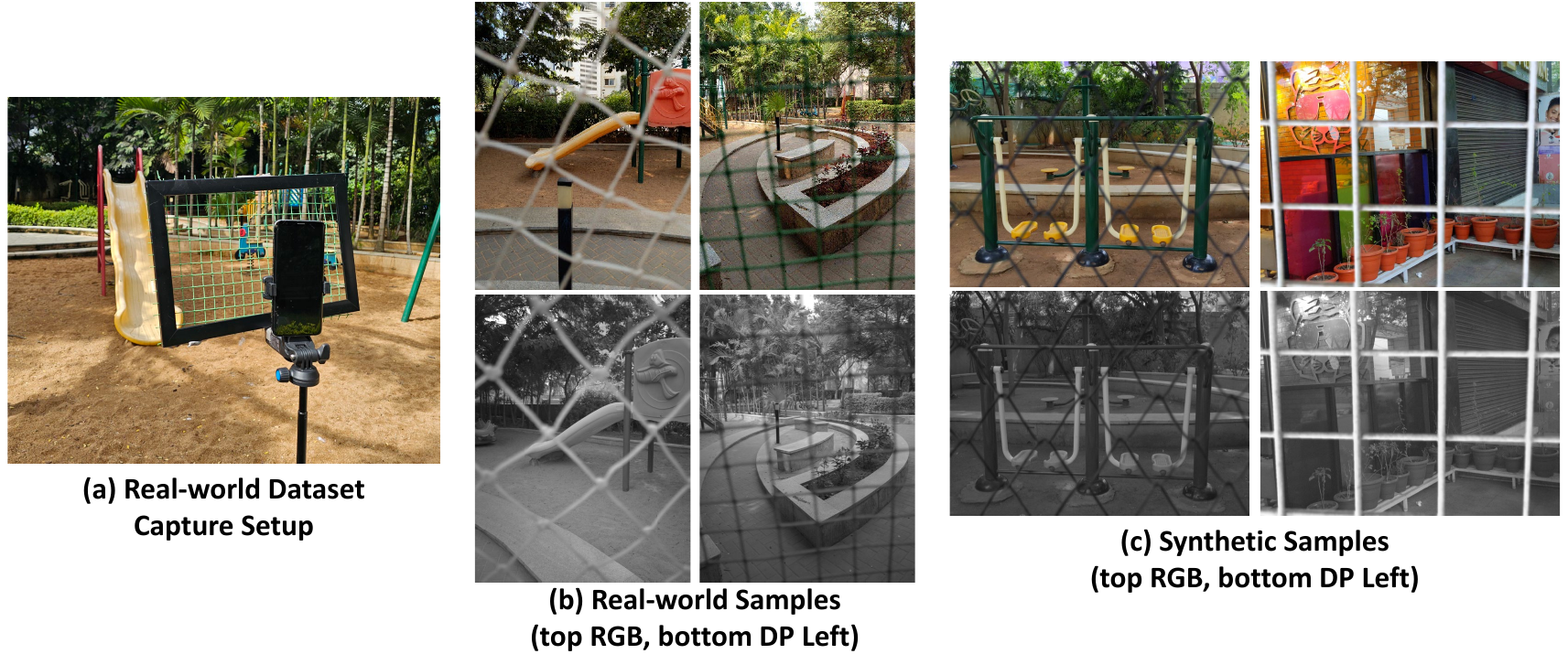}
	\vspace{-0.3cm}
    \caption{Real-world dataset capture setup, and real-world and synthetic dataset samples. Please zoom in for better detail.}
	\label{fig:realandsyntheticdatasetsamples}
    \vspace{-0.7cm}
\end{figure}

\vspace{-0.3cm}
\subsection{Real-World Dataset Collection}
\label{subsec:realworlddataset}
\vspace{-0.15cm}

To create a benchmark for real-world evaluation, we captured $150$ paired scenes. Using a tripod-mounted Google Pixel $4$ with its focus set on the background, we recorded a clean scene, followed by an occluded version using custom-built fence frames. To ensure diversity, these frames featured the various physical fences (some are shown in Fig.~\ref{fig:diversityoffences}) which were placed at varying distances from the camera to produce a range of defocus effects. Raw DP views were extracted using a custom application based on \cite{dp_depth_iccv2019,dp_depth_iccv2019_android_app}. Finally, to ensure pair quality, all images were manually checked for major misalignments and normalized for lighting via histogram matching. Our capture setup and samples are detailed in Fig.~\ref{fig:realandsyntheticdatasetsamples}.

\begin{figure}[t!]
    \vspace{-0.2cm}
    \begin{algorithm}[H]
    \caption{DP and RGB Fence Synthesis Pipeline}
    \label{alg:synthesis}
    \begin{algorithmic}[1]
    \STATE \textbf{Input:} Clean views ($I_L, I_R, I_C$), Fence RGB texture $F_{rgb}$, Mask $M$, PSF grids ($K_L, K_R$).
    \STATE \textbf{Output:} Occluded views ($I'_L, I'_R, I'_C$).
    \STATE $K_C \leftarrow (K_L + K_R) / 2$
    \STATE $d \sim U(10\text{cm}, 50\text{cm})$; $\alpha \propto 1/d$
    \STATE $K_v^{\alpha} \leftarrow \text{scale}(K_v, \alpha)$ for $v \in \{L, R, C\}$.
    \STATE Apply identical geometric augmentations to $F_{rgb}, M$.
    \STATE Apply color augmentations only to fence texture $F_{rgb}$.
    \STATE $F \leftarrow F_{rgb\text{'s green channel}}$ \COMMENT{Use green channel for DP}
    \FOR{$v$ in $\{L, R, C\}$}
        \IF{$v=C$}
            \STATE Process each channel $I_{C,c}$ with $F_{rgb,c}$ independently.
        \ELSE
            \STATE Process single-channel view $I_v$ with grayscale fence $F$.
        \ENDIF
        \STATE $I_{v,sharp} \leftarrow I_v \odot (1-M) + F \odot M$.
        \STATE $I_{v,blur} \leftarrow \text{PatchwiseConv}(I_{v,sharp}, K_v^{\alpha})$.
        \STATE $M_{v,blur} \leftarrow \text{PatchwiseConv}(M, K_v^{\alpha})$.
        \STATE $I'_{v} \leftarrow I_v \odot (1-M_{v,blur}) + I_{v,blur} \odot M_{v,blur}$.
    \ENDFOR
    \STATE \textbf{return} ($I'_L, I'_R, I'_C$).
    \end{algorithmic}
    \end{algorithm}
    \vspace{-1cm}
\end{figure}

\vspace{-0.3cm}
\subsection{Synthetic Dataset Generation}
\label{subsec:syntheticdataset}
\vspace{-0.1cm}

To enable robust, physics based training, our synthesis pipeline composites new, all-in-focus fence assets onto real, clean DP and RGB backgrounds. This hybrid approach preserves the sensor's authentic optical properties (e.g., lens vignetting, sensor noise). As existing datasets \cite{video_defencing_icme2018} lack diversity and provide pre-blurred fences unsuitable for our renderer, we captured and manually segmented various all-in-focus fences, some visual examples of which are shown in Fig.~\ref{fig:diversityoffences}. To further increase diversity, we augment these assets before compositing: geometric augmentations are applied to both the fence texture $F$ and its mask $M$ for alignment, while color augmentations are applied only to the texture $F$.

Our simulation leverages the spatially-varying PSFs calibrated by \cite{dp_modelling_iccv2021}, which require a patch-wise application. To simulate varying fence distances, we randomly sample a fence depth $d$ and calculate the corresponding blur scale $\alpha$ based on the thin lens model ($\alpha \propto |\frac{1}{d_{focus}} - \frac{1}{d}|$, where $d_{focus}$ is the focus distance) \cite{multipleviewgeometry_zisserman2003}. To model a camera focused on a distant background, we assume an infinite focus distance, which simplifies the relationship to $\alpha \propto 1/d$. The full synthesis pipeline, which involves compositing an augmented fence onto the clean views, blurring both the composite and mask, and then performing a final blend, is detailed in Alg.~\ref{alg:synthesis}. This process is applied to the DP views and to each channel of the RGB view independently. The final dataset consists of $904$ synthetic samples ($804$ train, $100$ test). Following \cite{dpdnet_eccv2020}, from the high-resolution ($2016 \times 1536$) training samples, we extract $\approx13.7$k patches of size $512 \times 512$ for training.

\vspace{-0.2cm}
\section{Proposed Method}
\label{sec:method}

\vspace{-0.2cm}
\subsection{Freq-DP Net}
\begin{figure}[t!]
	\centering
	\includegraphics[scale=0.465]{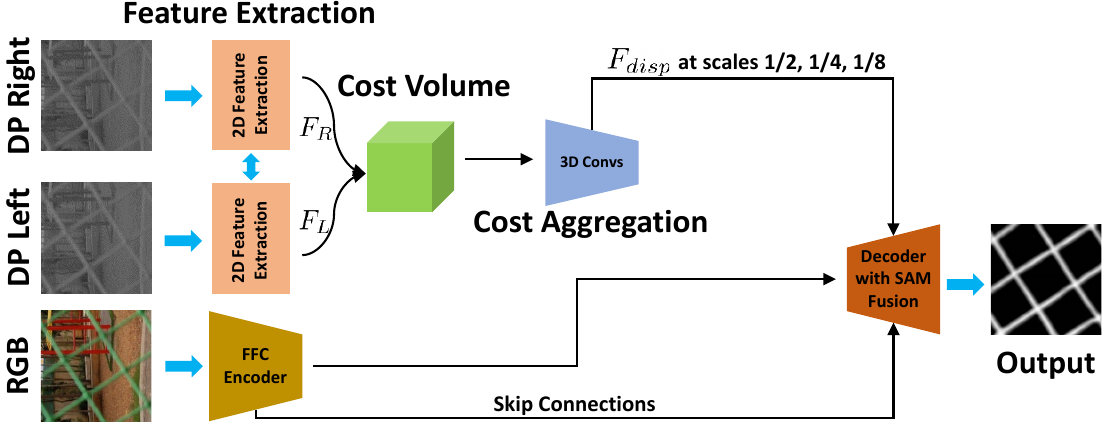}
	\vspace{-0.55cm}
    \caption{Network architecture of Freq-DP Net.}
	\label{fig:proposedmethod}
    \vspace{-0.55cm}
\end{figure}

To accurately segment the fence, we propose Freq-DP Net, a dual-branch encoder-decoder network that fuses geometric disparity cues with structural frequency-domain priors. The overall architecture is shown in Fig.~\ref{fig:proposedmethod}.

\vspace{-0.3cm}
\subsubsection{DP Disparity Branch}

This branch takes the DP views ($I_L, I_R$) as input. A shared-weight $2$D feature extractor based on residual blocks first produces feature maps $F_L, F_R \in \mathbb{R}^{C' \times H/2 \times W/2}$. We then construct a $3$D cost volume \cite{psmnet_cvpr2018} by correlating these features at sub-pixel precision:

\vspace{-0.2cm}
\begin{equation}
    C(\mathbf{p}, d) = \langle F_L(\mathbf{p}), \phi(F_R, \mathbf{p} - \mathbf{d}) \rangle
    \label{eq:cost_volume}
\end{equation}

where $\mathbf{p}=(x,y)$ is a pixel coordinate, $\mathbf{d}=(d,0)$, $\langle \cdot, \cdot \rangle$ is the inner product, and $\phi$ is the phase-shift interpolation \cite{light_field_depth_cvpr2015} function. Since the fence lies on one side of the focal plane, the DP disparity is unidirectional \cite{synthesize_dp_iccp2023}, defining our search range as $d \in [0, D_{max}]$. This volume is processed by a series of $3$D convolutions to produce disparity features ($F_{disp}$) at scales $H/2 \times W/2$, $H/4 \times W/4$, and $H/8 \times W/8$.

\vspace{-0.25cm}
\subsubsection{Structural Frequency Branch \& Fusion}

The structural branch is a symmetric, U-Net \cite{unet_miccai2015} like encoder-decoder built with FFC convolution layers \cite{chi2020fast} that processes the composite (RGB) image $I_C$. To leverage both cues, the disparity features $F_{disp}$ from the DP branch are fused into the FFC decoder. This fusion is performed selectively by a Structural Attention Module (SAM) only at the stages where disparity features are available ($H/2 \times W/2$, $H/4 \times W/4$, and $H/8 \times W/8$). At these stages, $F_{disp}$ modulates the FFC decoder's features $F_{ffc}$:

\vspace{-0.3cm}
\begin{equation}
    F'_{ffc} = F_{ffc} \odot \sigma(\text{Conv}_{1\times1}(F_{disp}))
    \label{eq:sam_fusion}
\end{equation}

where $F'_{ffc}$ are the refined features passed to the next decoder block. This guides the decoder to focus on regions that exhibit both a strong disparity signal and a fence-like structure. At other scales, the FFC decoder relies solely on its own skip connections. The final layer of the FFC decoder produces the soft segmentation mask.

\vspace{-0.2cm}
\subsection{Fence Removal}

For the final restoration, the predicted mask and RGB image are fed to LaMa-Fence, our version of the LaMa \cite{lama_wacv2022} inpainting network finetuned on our synthetic data to handle thin, repetitive fence patterns.

\vspace{-0.1cm}
\section{Results and Discussion}
\label{sec:resultsanddiscussion}
\vspace{-0.2cm}

We train Freq-DP Net on our synthetic dataset of size $\approx13.7$k and evaluate it on two test sets: $150$ real-world scenes and a held-out synthetic set of $100$ scenes, which also include unseen fence types to test for generalization. We assess performance on both the primary task of fence segmentation and the final application of fence removal.

\vspace{-0.2cm}
\subsection{Fence Segmentation}
\vspace{-0.2cm}

\begin{figure}[t!]
	\centering
	\includegraphics[scale=0.43]{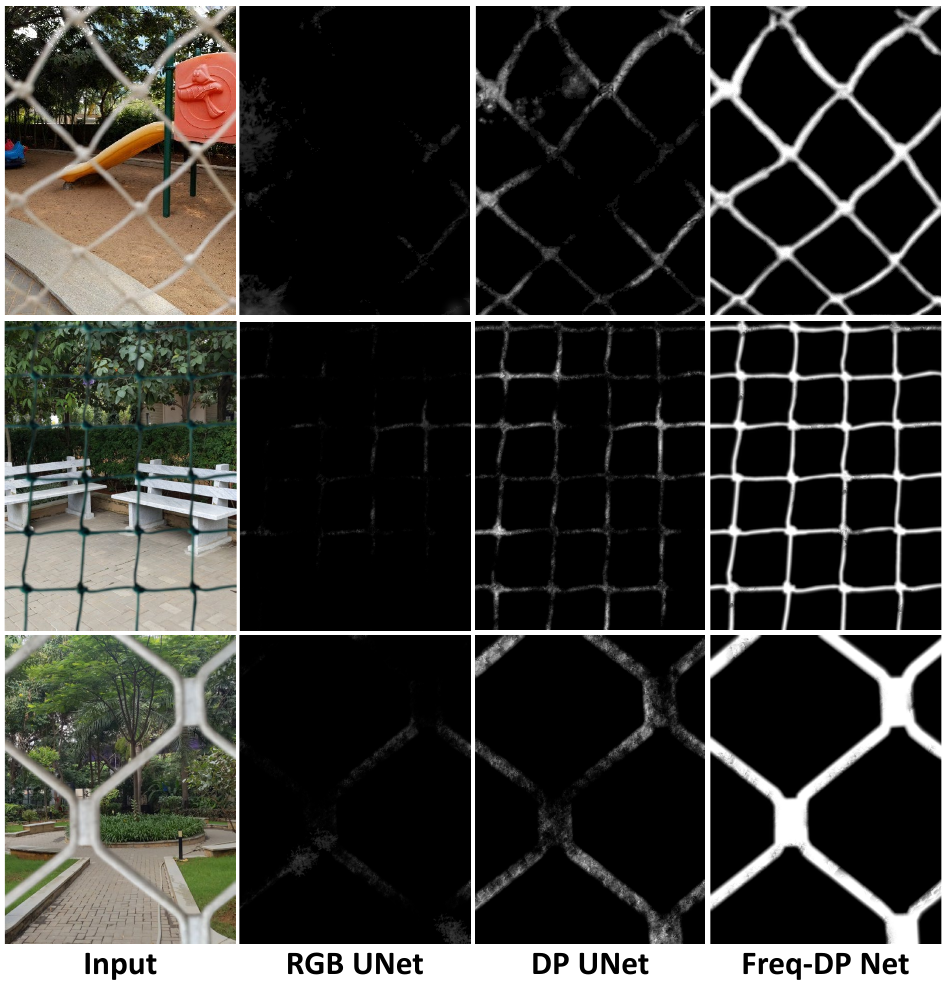}
	\vspace{-0.35cm}
    \caption{Qualitative results for fence segmentation. The third row highlights our model's successful generalization to a fence type unseen during training.}
	\label{fig:qualitativeresultsfencesegmentation}
    \vspace{-0.3cm}
\end{figure}

We evaluate our core contribution of fence segmentation against baselines with a similar parameter count for a fair comparison. As shown in Tab.~\ref{tab:quantitativecomparisonfencesegmentation}, our Freq-DP Net significantly outperforms both RGB U-Net and DP U-Net (RGB+DP concatenated input). The inferior performance of the RGB U-Net confirms that a single RGB modality is insufficient for this diverse task. Freq-DP Net's success is mirrored in the qualitative results in Fig.~\ref{fig:qualitativeresultsfencesegmentation}, where our method produces cleaner masks and demonstrates superior generalization to challenging, out-of-distribution fence types (e.g., the third row). This validates our physics-based, dual-branch design, which is more robust than appearance-based models.

\begin{table}[t!]
	\caption{Quantitative comparison for fence segmentation.}
	\label{tab:quantitativecomparisonfencesegmentation}
	\vspace{-0.3cm}
	\centering
	\resizebox{0.27\textwidth}{!}{
		\begin{tabular}{l|c|c|c}
			\hline
			\textbf{Method} &              \textbf{Precision}  & \textbf{Recall}   & \textbf{F1} 		   \\
			\hline
            \textbf{RGB UNet}        & 	0.776 	            & 	0.753 	        & 	0.764 	               \\
            \textbf{DP UNet} 	     &	0.936 	            & 	0.921 	        & 	0.928		           \\
            \textbf{Freq-DP Net} 	 &	\textbf{0.979}	    &  	\textbf{0.964} 	&	\textbf{0.971}	       \\
			\hline
		\end{tabular}
	}
	\vspace{-0.3cm}
\end{table}

\begin{table}[t!]
    \caption{Quantitative comparison for end-to-end fence removal on our synthetic and real test sets.}
    \label{tab:quantitativecomparisonfenceremoval}
    \vspace{-0.3cm}
    \centering
    \resizebox{0.48\textwidth}{!}{
        \begin{tabular}{lcc|cc}
        \toprule
        \multirow{2}{*}{\textbf{Method}} & \multicolumn{2}{c|}{\emph{Synthetic Test Set}} & \multicolumn{2}{c}{\emph{Real Test Set}}       \\
        \cmidrule(lr){2-3} \cmidrule(lr){4-5}
         & \textbf{PSNR $\uparrow$} & \textbf{SSIM $\uparrow$} & \textbf{PSNR $\uparrow$} & \textbf{SSIM $\uparrow$}                       \\
        \midrule
        \textbf{Restormer (RGB)}                       & 32.23          & 0.9457             & 29.78            & 0.8875                   \\
        \textbf{LaMa (RGB)}                            & 30.67          & 0.9323             & 28.45            & 0.8763                   \\
        \textbf{Restormer (RGB+DP)}                    & 39.89          & 0.9754             & 31.54            & 0.9012                   \\
        \textbf{LaMa (RGB+DP)}                         & 37.72          & 0.9623             & 30.67            & 0.8932                   \\
        \textbf{Ours (Freq-DP Net + LaMa-Fence)}       & \textbf{41.28} & \textbf{0.9876}    & \textbf{34.54}   & \textbf{0.9223}          \\
        \bottomrule
        \end{tabular}
    }
    \vspace{-0.6cm}
\end{table}

\vspace{-0.4cm}
\subsection{Fence Removal}
\begin{figure}[t!]
	\centering
	\includegraphics[scale=0.37]{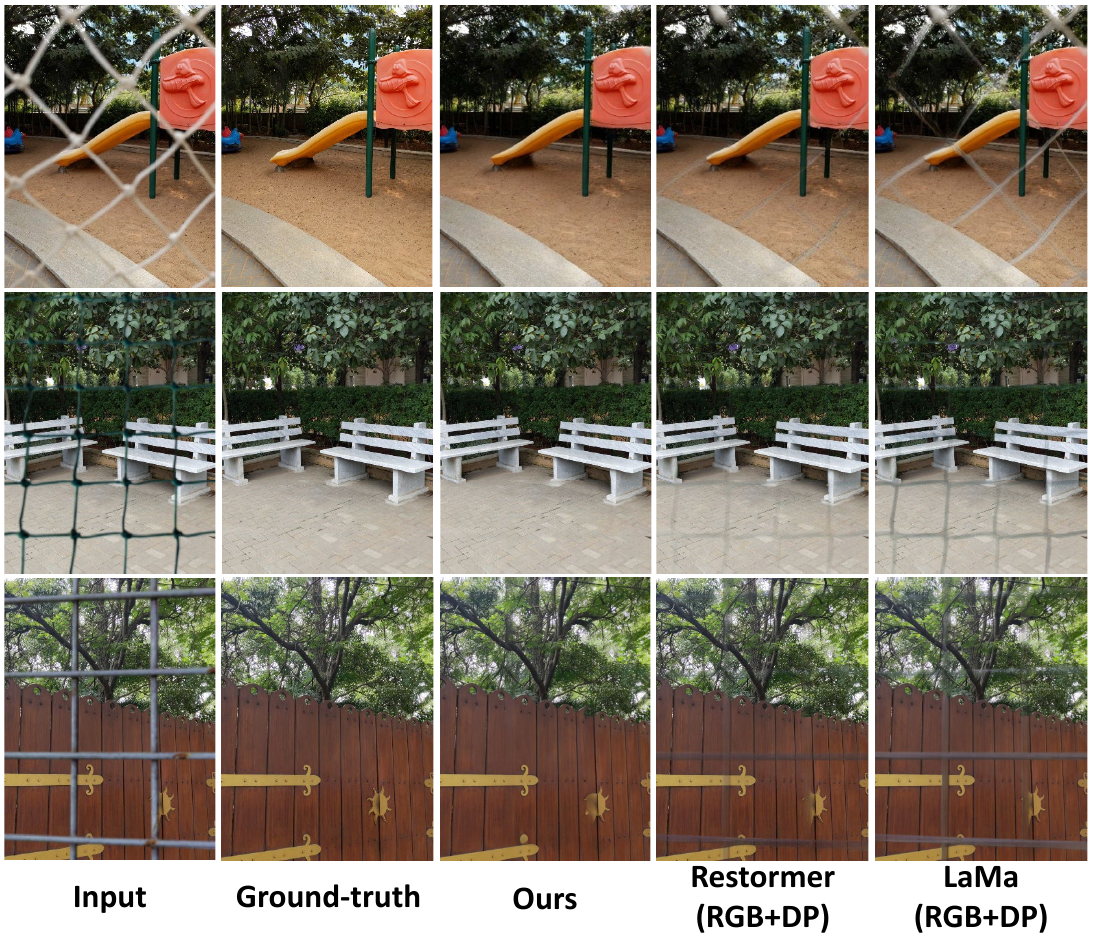}
	\vspace{-0.37cm}
    \caption{Qualitative results for fence removal.}
	\label{fig:qualitativeresultsfenceremoval}
    \vspace{-0.3cm}
\end{figure}

\vspace{-0.2cm}
Next, we evaluate the final restoration quality. We compare our two-stage pipeline against powerful end-to-end baselines: Restormer \cite{restormer_cvpr2022} and LaMa \cite{lama_wacv2022}. We test two variants for each baseline: an RGB-only version, and an RGB+DP version that takes a 5-channel input from the concatenated RGB and DP images. Tab.~\ref{tab:quantitativecomparisonfenceremoval} shows our method achieves higher PSNR and SSIM scores on both synthetic and real-world test sets. The strong performance on the real-world set is particularly important, as it demonstrates our model's effective generalization from synthetic training data. Visual comparisons in Fig.~\ref{fig:qualitativeresultsfenceremoval} also show that our approach performs more accurate restoration with fewer artifacts. Note that we only visualize the stronger performing RGB+DP variants of the baselines.

\vspace{-0.4cm}
\subsection{Ablation Study}
\vspace{-0.2cm}

Finally, our ablation study in Tab.~\ref{tab:ablationstudyfencesegmentation} quantitatively validates our architectural choices. \textcolor{black}{All ablation variants use same parameters as our full model ($\approx22.4$M) to control for model capacity.} The results clearly demonstrate the contribution of each component. Removing either the DP cost volume or the FFC encoder significantly degrades segmentation performance, confirming that explicitly modeling both geometric disparity and global structure are crucial. Furthermore, we test a variant where our SAM Fusion is replaced with simple feature concatenation. This also reduces performance, proving that the intelligent fusion via SAM is more effective than a na\"ive merge, and that all proposed components work synergistically to achieve the final result.

\begin{table}[t!]
	\caption{Ablation study for fence segmentation.}
	\label{tab:ablationstudyfencesegmentation}
	\vspace{-0.3cm}
	\centering
	\resizebox{0.35\textwidth}{!}{
		\begin{tabular}{l|c|c|c}
			\hline
			\textbf{Ablation Setting} & \textbf{Precision} & \textbf{Recall} & \textbf{F1} 			                              \\
			\hline
			\textbf{No Cost Volume (in short, CV)}             &    0.943	            &  	0.931	        &	0.937 		      \\ 
			\textbf{No FFC based Encoder (in short, FFC)} 	   &	0.958	            &  	0.947 	        &	0.952		      \\
            \textbf{No CV and FFC (same as DP UNet)} 	       &	0.936 	            & 	0.921 	        & 	0.928		      \\
            \textbf{No SAM Fusion} 	                           &	0.968	            &  	0.955	        &	0.961		      \\
            \textbf{Freq-DP Net} 	                           &	\textbf{0.979}	    &  	\textbf{0.964} 	&	\textbf{0.971}	  \\
			\hline
		\end{tabular}
	}
	\vspace{-0.47cm}
\end{table}

\vspace{-0.25cm}
\section{Conclusion and Future Work}
\label{sec:conclusion}
\vspace{-0.3cm}

We introduced the first framework leveraging dual-pixel (DP) sensors for single-image fence removal. \textcolor{black}{Our network, Freq-DP Net, establishes a new state-of-the-art on our newly released diverse benchmark. Despite its performance, our method has limitations: DP cues weaken as fence distance increases, and the single-image formulation requires generative inpainting for occluded regions, whereas video-based methods can retrieve pixels from adjacent frames for more faithful results. Nevertheless, this work demonstrates the power of physics-based sensor cues for complex occlusions.} Future work will handle multiple occlusion types in a unified model.

\bibliographystyle{IEEEbib}
\bibliography{references}

\end{document}